\pdfoutput=1

\documentclass[11pt]{article}

\usepackage[]{ACL2023}

\usepackage{times}
\usepackage{latexsym}
\usepackage{graphicx}
\usepackage{multirow}
\usepackage{float}

\usepackage[T1]{fontenc}

\usepackage[utf8]{inputenc}

\usepackage{microtype}
\usepackage{array}
\usepackage{arydshln}
\usepackage{inconsolata}

\title{FS-RAG: A Frame Semantics Based Approach for Improved Factual Accuracy in Large Language Models}

\author{Harish Tayyar Madabushi \\
  The University of Bath, Bath, UK\\
  \texttt{htm43@bath.ac.uk} \\
  }

\begin{document}
\maketitle
\begin{abstract}
We present a novel extension to Retrieval Augmented Generation with the goal of mitigating factual inaccuracies in the output of large language models. Specifically, our method draws on the cognitive linguistic theory of frame semantics for the indexing and retrieval of factual information relevant to helping large language models answer queries. We conduct experiments to demonstrate the effectiveness of this method both in terms of retrieval effectiveness and in terms of the relevance of the frames and frame relations automatically generated. Our results show that this novel mechanism of Frame Semantic-based retrieval, designed to improve Retrieval Augmented Generation (FS-RAG), is effective and offers potential for providing data-driven insights into frame semantics theory. We provide open access to our program code and prompts\footnote{\href{https://github.com/H-TayyarMadabushi/A-Frame-Semantics-based-approach-for-Improved-Factual-Accuracy-in-Large-Language-Models}{https://github.com/H-TayyarMadabushi/A-Frame-Semantics-based-approach-for-Improved-Factual-Accuracy-in-Large-Language-Models}}.
\end{abstract}

\section{Introduction, Motivation and Context}
Large language models (LLMs), despite their significant capabilities and widespread adoption have the inherent tendency to generate plausible sounding, yet inaccurate, output. This phenomenon, referred to as ``hallucinations,'' has been a significant stumbling block in the widespread deployment of these solutions~\cite{10.1145/3571730}. Hallucinations themselves are not limited to factual inaccuracies, and include other modes of failure, including the incorrect interpretation of input prompts and errors in logical inference. However, factual hallucinations are particularly important to address as the retrieval of incorrect facts is particularly hard to recover from and can neutralise and make irrelevant all other improvements. 

The capabilities of LLMs typically improve with an increase in their ``size,'' which is a combination of a model's parameters and the size of the pre-training corpus. Until recently, this was seen by some as being evidence that further scaling would eventually address the shortcomings of LLMs, including hallucinations. For example, LLMs were claimed to develop ``emergent abilities'': specifically, it was believed that LLMs, when scaled to several billion parameters developed capabilities including those required to solve tasks requiring reasoning in humans, thus indicative that they were developing reasoning skills~\cite{wei2022emergent}. More recent work, however, has shown that this is not the case and that LLMs instead develop a single capability, which they leverage to solve tasks~\cite{lu2023emergent}. This capability, called ``in-context learning,'' is, roughly put, the ability of models to solve a particular task based on a few examples provided in the prompt~\cite{NEURIPS2020_1457c0d6,JMLR:v24:22-1144}. \citep{lu2023emergent} further suggest that the process of instructional fine-tuning LLMs to understand instructions~\cite{wei2022finetuned}, enables models to leverage the same ``in-context'' abilities even in the absence of examples. This finding indicates that further scaling, while providing improved instruction following abilities, will not grant modes the broader capacity to for general reasoning.

The fact that LLMs are not likely to develop the ability to reason has profound implications to work on improving them, including to mitigating hallucinations. It implies that we must explore alternative approaches. This is especially the case when it comes to factual hallucinations as the `parametric memory' in LLMs is orders of magnitude smaller than the pre-training data~\cite{10.1145/3571730}. As such, they must necessarily use some method of compressing their pre-training data. Without the ability to distinguish between the information that is relevant and what is not relevant in their pre-training data, their method of compression defaults to be the memorisation of frequent information and the learning of statistical patterns and trends to represent the less frequent information. 

\subsection{Retrieval Augmented Generation}
Given this context, it isn't surprising that Retrieval Augmented Generation, or RAG, which involves the inclusion of relevant information to the prompt has been so successful~\cite{NEURIPS2020_6b493230}. It provides a viable mechanism of offloading information retrieval (IR) demands and instead focuses on using the LLM as a mechanism of analysing and processing data based on explicit instructions. It should be noted that analysing and processing data through explicit instructions is precisely what models excel at through in-context learning. Importantly, RAG also provides a solution to another of LLMs' problems, which is the outdating of knowledge. By incorporating the latest information, RAG prevents the need for further training for even minor knowledge updates to the LLM which is clearly infeasible~\cite{shuster-etal-2021-retrieval-augmentation}. 

However, RAG comes with its own shortcomings. RAG transfers the problem of mitigating factual hallucinations to one of retrieving information relevant to answering a query~\cite{gao2024retrievalaugmented}. While LLMs can handle some noise in the retrieved context provided, a dramatic increase in noise unsurprisingly leads to deteriorating performance of models. The retrieval of information relevant to answering a query is non-trivial as such information is not always likely to be semantically related to the query. This problem becomes even more important when the query requires reasoning over multiple facts each of which are progressively semantically further from the query. Overall, the fact that logically connected information is not always semantically similar makes existing keyword and semantic similarity based search and IR systems poorly suited for the specific IR requirements of LLMs. 

Existing mechanisms of dealing with this problem in information retrieval typically involve using LLMs themselves to solve this problem. Broadly, there are two ways of doing this: the first involves the use of LLMs to decompose the query into sub-queries each of which require less complex reasoning~\cite{patel-etal-2022-question,zhou2023leasttomost}, and the second involves the use of LLMs to generate intermediary reasoning steps, called chain of thought~\cite{wei2022chain}, to answer the query. These intermediary steps are then used to generate IR search queries and the resultant facts are fed back to an LLM to generate a response~\cite{shao-etal-2023-enhancing}. While these methods show some promise, the problem with them is clear: they are yet another opaque mechanism the failure of which will be hard to fix. In fact, we might be able to improve these systems be fine-tuning models to perform this task. However, such a solution is only going to further obfuscate the process. 

\subsection{Contributions}
\emph{To address these concerns we propose a novel, transparent, and mutable storage and retrieval system for the mitigation of factual hallucinations in LLMs. This solution is based on theoretical insights from cognitive linguistics, specifically frame semantics.} Specifically, this work makes the following contributions: 
\begin{enumerate}
    \item We propose a novel, transparent, and mutable storage and retrieval system for the mitigation of factual hallucinations in LLMs which is based on theoretical insights from frame semantics. We call this system \textsc{FS-RAG}.
    \item We show the effectiveness of our method by experimenting on a closed domain question answering. Specifically, our method of storage and retrieval provides a significant boost over both treditional search systems and, significantly, also outperforms an LLM based search system wherein we generate search terms using a Language Model.
    \item We show that our method has the significant additional advantage of being interpretable, and thus having the potential to provide data-driven insights to the theory of frame semantics. 
\end{enumerate}

\section{Frame Semantics}
Given the context provided earlier, it is evident that only the most frequently occurring factual information from pre-training is explicitly retained in LLMs. Less frequently occurring facts are not explicitly stored and instead the model has access to only statistical approximations. Given that the exact information stored is not explicit and also different for models of different scale and training regimes, the only way to get around hallucination is to explicitly provide LLMs with all but the most common information The effectiveness of RAG provides an easy mechanism for including such information. However, as mentioned previously, the effectiveness of this method hinges on the non-trivial task of retrieving  relevant information.

We propose that this mechanism can be found in the concept of Frame Semantics in cognitive linguistics, which purportedly facilitate human understanding of words by allowing us to recall pertinent information. 

Frame semantics~\cite{fillmore2006frame} is a theory of linguistic meaning that emphasises that the meanings of words are best understood by the semantic and conceptual ``frames'' or ``schemas'' within which they function. A frame is a cognitive structure that helps individuals understand and perceive the world around them, enabling them to organise knowledge based on typical situations, actions, or common experiences. In linguistics, a frame influences how the meanings of words are interpreted in different contexts. For example, the word ``sell'' invokes a commercial transaction frame involving a seller, a buyer, an item being sold, and an agreed price and helps to predict and explain the use of other related words and the roles they play within the same context. FrameNet \cite{baker-etal-1998-berkeley-framenet} is an online database based on frame semantics, with the goal to catalogue English words and their associated semantic frames, defining the various roles and relations in a frame and illustrating these with example sentences. Each ``frame'' in FrameNet captures a specific type of event, relation, or entity and the roles associated with it. For example, the \textsc{commerce\_sell} frame in FrameNet includes roles for the seller, the buyer, the goods being sold, and is annotated to be inherited by the frame \textsc{renting\_out} and as a ``perspective on'' the frame \textsc{commerce\_goods-transfer}.

\section{Frame Semantic RAG}
\label{sec:fsrag}
This section provides an overview of Frame Semantic RAG (FS-RAG), the proposed mechanism of storing and indexing factual information to aid effective retrieval for the mutation of hallucinations in LLMs. In evaluating our mechanism of retrieval we make use of Entailment Bank, \cite{dalvi-etal-2021-explaining}, which comprises science questions from school years 4 to 6, along with relevant facts and ``entailment trees''. 

\begin{table}[h!]
\renewcommand{\arraystretch}{1.5}
\begin{center}
\begin{tabular}{ | m{1.8cm} | m{5cm}| }
\hline
Question & How might eruptions affect plants?  \\ 
\hdashline
Associated Facts & F1: eruptions emit lava; \newline F2: eruptions produce ash clouds; \newline F3: plants have green leaves; \newline F4: plant producers die without sunlight; \newline F5: ash clouds block sunlight.  \\
\hdashline
Inference Steps &  F2 + F5 implies I1: eruptions block sunlight; \newline
F4 + I1 implies I2: eruptions can cause plants to die. \\
\hdashline
Answer &  eruptions can cause plants to die. \\
\hline
\end{tabular}
\caption{\label{table:ebexample}An example question from Entailment Bank and associated factoids. Language models find it significantly easier to generate the required entailment trees when presented with all relevant facts. However, the fact that the retrieval of relevant facts is non-trivial motivates a frame semantics based mechanism for indexing and search facts relevant to answering questions.}
\end{center}
\end{table}

Consider Table \ref{table:ebexample}, which presents an example from the Entailment-Bank dataset. The original task involves building an entailment tree to answer questions and consists of three tasks at different levels of difficulty: a) Task 1 presents the model with all relevant facts and requires only the construction of the entailment tree; b) Task 2 requires the model to perform the same task, but with 15 to 25 distractors included; c) Task 3 involves first extracting the relevant facts from the entire corpus before constructing the entailment tree. The authors find that even a relatively small model, T5-11B \cite{10.5555/3455716.3455856}, can perform relatively well on Tasks 1 and 2, when fine-tuned. Task 3, they find, is much harder highlighting the importance of efficient retrieval. We direct the reader to the results section of \citep{dalvi-etal-2021-explaining} for details.

Overall, these results reinforce our earlier points: Retrieval is non-trivial and improving retrieval has the potential to significantly boost model performance. In the example presented above,  using search terms derived just from the question (e.g. "eruption") including more complex combinations (e.g. ``eruption and plants'') may not effectively retrieve relevant information. Additionally, if the search terms are too broad, it can cause the retrieval of a significant number of irrelevant facts. Both the lack of relevant facts and a large number of unrelated facts can hinder the model's performance. 

\emph{This work is motivated by the hypothesis that we can significantly narrow the search space if we index facts--stored as plain text--according to the frames they invoke and use the frames associated with the question along with the relations between frames to retrieve relevant facts.} To test our hypothesis, we focus our experiments on the retrieval of relevant facts. 

In the above example, the FrameNet frames associated with the question could include: a) \textsc{Surviving}, which captures situations requiring endurance in a dangerous situation and includes annotations of the frame element ``Dangerous\_situation''; and b) \textsc{Cause\_harm}, which describes a situation where an `agent' injures a `victim'. Our frame-based mechanism additionally allows the exploitation of the relations between frames to traverse the hierarchical interconnections between frames. This method approximates reasoning steps, enabling the retrieval of facts that are logically connected, even if they are not semantically similar. Within the limited scope of a single task, this work shows a significant improvement in retrieval (recall) using frames when compared to traditional search based baseline, and when compared to retrieval using search terms generated by LLMs, thus verifying the feasibility of this approach. Finally, we use the adaptability of state of the art LLMs to develop helper-LLMs that generate sets of frames, which we then evaluate and compare for their effectiveness in retrieving relevant information. Expanding these methods has the potential to provide, for the first time, data-driven insights that can be used to refine the theory of frame semantics.

\section{Experimental Settings}
In this section, we describe the experiments we conduct to test the hypothesis that frame based indexing and search is a more effective mechanism than keyword based indexing. 

\subsection{Task}
Our choice of the specific task is motivated by the observation that LLMs can perform reasonably well when provided with relevant facts alongside some distractors. However, as described in the previous section, the retrieval of these relevant facts poses a significant challenge. Therefore, we focus on the task of retrieving relevant factoids for answering questions in Entailment Bank. Specifically, we focus on the information extraction subtask required in Task 3 described in Section \ref{sec:fsrag}. Notice that the effective retrieval of facts would simplify Task 3 to Task 2, the task of building entailment trees given the relevant facts and some distractors. Given how effective T5-11B, which, by current standards consists of relatively few parameters, is on Task 2, simplifying Task 3 to Task 2 provides a template for solving tasks based exclusively on retrieved facts, which would in turn help with the mitigation of factual hallucinations in LLMs. We slightly modify Task 3 by constructing the corpus of facts that we extract from using all the facts required by any question across the relevant data split, instead of the complete text book corpus which is harder to process. Regardless, we evaluate frame semantic retrieval and the baselines on exactly the same set of questions and facts to ensure a fair comparison. 

\subsection{Empirical Evaluation Metrics}
Given the nature of our task, we select Retrieval@k as our evaluation metric. The average length of  entailment trees in the Entailment Bank dataset 7.6 with very few having more than 10. Given that Task 2 (described previously in Section \ref{sec:fsrag}) includes between 15 and 25 distractors, we test our methods using Retrieval@k for $k \in 35, 40, 45$. Success in this setting will demonstrate that our retrieval mechanism can effectively simplify Task 3, which requires retrieval from the entire corpus, into Task 2, which involves building entailment trees based on relevant facts and a few distractors. Recall that models perform significantly better on Task 2 than on Task 3.

\subsection{Baselines}
\label{sec:baseline}
We use two different baselines to compare the effectiveness of frame semantic indexing and retrieval against. The first, is a simple keyword match baseline and is chosen due to our emphasis on interpretability and ease of correction. Being able to understand why a system retrieves certain facts enables refinement of the system by reassigning facts to different clusters or index buckets. This level of transparency is not possible with more opaque methods, such as dense vector-based retrieval. Since frame semantic retrieval implicitly provides interpretable we choose a baseline that is similarly transparent. We first generate search terms by feeding the relevant question to RAKE~\cite{rose2010automatic}, a tool for extracting search terms, which is known to be effective. We then perform a simple string match to extract all factoids that contain the keywords.  

The second baseline we use is not directly comparable as it is not interpretable. This consists of using an LLM to generate relevant search terms and is included so we might compare frame semantic retrieval to a rough analogue of question decomposition, a standard mechanism used in RAG as described above. Both baselines can be boosted using several techniques. For example, we do not generate chain of thoughts before generating the search terms using GPT-4, which is likely to improve the effectiveness of the resultant search terms. This is because the purpose of this work is not create a mechanism that outperforms existing methods, but to establish the feasibility of the frame semantic indexing and retrieval process which has the advantage of being interpretable and being based on cognitive linguistic theory. 

\section{Frame Semantic Retrieval: Methods and Qualitative Analysis}

\begin{table*}[h!]
\renewcommand{\arraystretch}{1.5}
\footnotesize
\begin{center}
\begin{tabular}{ | m{3cm} | m{5.6cm} | m{5.6cm} | }
\hline
\textbf{Task} & \textbf{Prompt} & \textbf{Output Example} \\
\hline
Frame Identification \newline\newline Facts are indexed by the the frames they invoke \newline During Inference, relevant facts are extracted based on frames invoked by the question and additional frames that are related.
& 
What is the single/two most important frame, based on the theory of frame semantics, relevant for answering the question/fact below. Do not include frames about answering questions or reasoning, that is implied. Do not include frames which are metaphorical. Ensure the the name of the frame is as descriptive as possible. Output a single frame and join words in the frame by underscores. Output nothing but the name of the frame. \newline 
Question 1:  How does the appearance of a constellation change during the night? 
\newline
Answer 1: celestial\_motion                                                                                  
\newline
\dots 
\newline
Problem:                                                                                                     \newline
Question Problem: <QUESTION>
\newline
Answer Problem: 
& Input Question: Tides, such as those along the coast of Massachusetts, are caused by gravitational attractions acting on Earth. Why is the gravitational attraction of the Moon a greater factor in determining tides than the gravitational attraction of the much larger Sun?
\newline
Output Frame: \textsc{gravitational\_influence}
\\
\hdashline
Check if the new frame must be added to the frame set \newline\newline Used during inference & 
The following question has been tagged with the single frame listed. Is this frame significantly different from existing frames listed and should it be added as a new frame? Respond with True if it is significantly different otherwise False. Respond with True and False only. 
\newline
Example Question: From Earth, the Sun appears brighter than any other star because the Sun is the
\newline
Example Tagged Frame:proximity 
\newline
Example Existing Frames 2: \textsc{celestial\_motion}
\newline
Example Answer: True
\newline
\dots 
\newline
Question Problem: <INPUT QUESTION>
\newline
Tagged Frame Problem: <INPUT NEW FRAME>                                                                      
\newline
Existing Frames Problem: <INPUT EXISTING FRAME>
\newline
Answer Problem:
&
Input Question:Melinda learned that days in some seasons have more daylight hours than in other seasons. Which season receives the most hours of sunlight in the Northern Hemisphere?
\newline
Input Frame Assigned: \textsc{seasonal\_variation\_in\_daylight}
\newline
Input List of Existing Frames: \textsc{daylight\_variation}, \textsc{seasonal\_adaptation}, \textsc{seasonal\_behavior}, \textsc{seasonal\_change}, \textsc{seasonal\_variation}
\newline
Output (Add \textsc{seasonal\_variation\_in\_daylight} to Frame Set?): False
\newline
Action Taken: Question tagged with \textsc{daylight\_variation}
\\
\hdashline
Identifying Frame Relations \newline\newline Used during inference &
Listed below is a single frame relevant to a question. List those frames which are most likely to be associated with the facts required for answer this question. These frames are  based on the theory of frame semantics. Do not include frames about answering questions or reasoning, that is implied. Do not include frames which are metaphorical. [\dots] 
\newline
Example Question 1: Stars are organized into patterns called constellations. One constellation is named Leo. Which statement best explains why Leo appears in different areas of the sky throughout the year?
\newline
Example Question Frame: \textsc{celestial\_motion}         
\newline
Example Output Frames: \textsc{constellation\_classification}, \textsc{star\_classification}, \textsc{celestial\_motion}
\newline 
Problem Question : <QUESTION>
\newline
Problem Question Frame :  <FRAME>
\newline
Problem Output Frames:
& 
Input Question: Which measurement is best expressed in light-years?
\newline
Input Question Frame:  \textsc{distance\_in\_astronomy}
\newline
Output set of Frames Related to Question Frame: \textsc{celestial\_distance}, \textsc{astronomical\_unit}, \textsc{spatial\_measurement}
\newline
\\
\hline
\end{tabular}
\caption{\label{table:prompts} Prompts and associated outputs for each step in frame based indexing and retrieval.}
\end{center}
\end{table*}

In this section we detail the methods used for frame semantic indexing and retrieval. Given that one of objectives is to maintain interpretability and to to potentially provide data-driven insights to the theory of frame semantics, we perform a qualitative analysis of the outputs of each of the stages. An empirical evaluation of the effectiveness of these methods is presented next, in Section \ref{sec:empresults}.

The mechanism of retrieving information based on frame semantics consists of three distinct stages: The first is a pre-processing step, which involves indexing all relevant factoids based on between two and four of the most prominent frames that they invoke. The second and third steps are performed when answering a question at inference time and involve identifying the single most important frame associated with the question (the question frame), and frames associated with the question frame, which are likely to be associated with factoids relevant to answering the original question but separated by one or more logical steps. The most important parts of the prompts associated with each of these steps along with example outputs are presented in Table \ref{table:prompts}. The complete prompts, including prior versions we experimented with, are included in the supplementary data uploaded with this paper. In all cases, we prompt \mbox{GPT-4}~\cite{openai2024gpt4} using a temperature of 0 to ensure reproducible results.

\subsection{Frame Identification}
There are two difficulties in identifying the frames associated with facts or questions. The first is the necessity to define a complete set of frames, and the second is the linking of these frames to the relevant fact or question. Our exploratory experiments of using FrameNet as a definitive source of all frames which we use to compare against facts and questions from Entailment Bank, showed that FrameNet is inadequate for this purpose. Specifically, the approximately 1,200 frames indexed on FrameNet have two significant shortcomings. The first is FrameNet's focus on `trigger' words to identify frames is problematic. This emphasis on individual trigger words, likely influenced by the tools available at the time of FrameNet's inception, overlooks the fact that a sentence, as a whole, might invoke a frame that is difficult to identify through trigger words alone, which themselves can be challenging to extract within sentences. The second is the fact that the frames available within FrameNet cover a limited set of domains, which overlap minimally with the frames that are appropriate for the Entailment Bank dataset. 

To address these issues, we bootstrap the creation of frames using an LLM, specifically \mbox{GPT-4}. We prompt \mbox{GPT-4} to generate frames relevant to the input fact or question, allowing us to organically expand our set of frames. We use in-context examples, selected from the training set, to enable the model to better output relevant frames. The complete prompts, including prior versions we experimented with, are included in the supplementary data uploaded with this paper. We start with an empty `frame set' and iteratively generate frames associated with facts and questions. For each fact or question, the frames output by \mbox{GPT-4} are compared with the existing frames previously generated (or none in the initial instances). This comparison is also done with the help of \mbox{GPT-4}. We first extract 5 frames, whose frame names are most semantically similar to that of the newly generated frame. This is done using sentence BERT~\cite{reimers-gurevych-2019-sentence}, an effective semantic similarity metric that originally relied on BERT~\cite{devlin-etal-2019-bert}, but now makes use of custom contextual embeddings. We then prompt \mbox{GPT-4} to determine if the newly generated frame must be added to the frame set. 

As an example, \mbox{GPT-4}, when prompted to generate frames related to the factoid  ``the gravitational pull of the sun on earth's oceans causes the tides,'' might generate \textsc{gravitational influence} and \textsc{tidal movement}. These frames are compared against the existing frames and the frame \textsc{gravitational influence} might be replaced by the similar frame \textsc{gravitational attraction} already in our frame set. If a similar frame is not found, the original frame is added to the frame set. This same process is then used to generate frames associated with questions. We find that this method is sub-optimal and that GPT-4 is a poor judge of identifying frames which are truly different form those already in the frame set. As such, we always augment the original set of frames with five frames whose names are most semantically similar to the original. See also Table \ref{table:prompts} for more examples.  

\subsection{Frame Relations}

\begin{table*}[ht]
\footnotesize
\begin{center}
\begin{tabular}{ | m{2cm} | m{2cm} | m{2cm} | m{4cm} | }
\hline
\textbf{Recall@} & \textbf{RAKE Search \newline (Baseline 1)} & \textbf{GPT-4 Search \newline (Baseline 2)}  & \textbf{Frame Semantic Retrieval \newline (our method)} \\
\hline
@35 & 0.330 & 0.385 & 0.439 \\
@40 & 0.333 & 0.390 & 0.464 \\
@45 & 0.338 & 0.396 & 0.473 \\
\hline
\end{tabular}
\caption{\label{table:empeval} Recall at k between 35 and 45 comparing frame semantic retrieval to search based retrieval where the search terms are generated using a traditional keyword based method (RAKE) and using GPT-4. It is notable that frame semantic retrieval performs significantly better than both baselines across all selected values of $k$.}
\end{center}
\end{table*}

We call the overlap between the frames invoked by a question and those invoked by the facts necessary for answering that question a first-order frame overlap. This first-order overlap is not sufficient to extracting all facts that are relevant to answering a question. As such, we require a means of identifying relations between frames, so we can expand the set of relevant frames to include those which are related to the original, as a proxy for the reasoning process. This is in addition to the expansion using semantic similarity described previously. 

Instead of importing definitions of frame relations, for example from FrameNet, we generate these relations using a data-driven approach. Specifically, we extract questions and associated facts from the training data. We then assign frames to both the questions and the facts using the methods described previously. The frames associated with the questions and the corresponding facts are assumed to hold a latent relation, which we use to generate similar frame relations at the time of answering questions. This is done by prompting GPT-4 with the relevant question and the frame associated with the frame and requiring GPT-4 to generate frames relevant to answering the question. Row three of Table \ref{table:prompts} presents the prompt and an example output of this step.

\subsection{Qualitative Analysis}
A qualitative analysis of resultant frames and frame relations demonstrate the surprising effectiveness of this method. Table \ref{table:prompts} presents some of the frames and frame relations automatically generated using the methods described above. The results are far from perfect, but are interesting from the perspectives of the diversity and adaptability they present. It is important to note that these results are achieved though prompting alone. Given that LLMs, such as GPT-4, are unlikely to be designed to solve tasks such as this, it is not surprising that there is much room for improvement, although the results demonstrate the feasibility of this method. Overall, we believe that bootstrapping these methods--by first manually refining a dataset generated through prompting, then iteratively training models specifically for this task, and subsequently using that model to generate more data--is an effective way to scale these methods.

\section{Empirical Evaluation}
\label{sec:empresults}
In addition to the qualitative analysis we present an empirical evaluation of the frame semantic retrieval methods described above. We compare the performance of frame semantic retrieval to the two search based baselines described in Section \ref{sec:baseline}. We present the results in Table \ref{table:empeval}. Overall, we find that frame semantic retrieval outperforms both the simple search based baseline and the baseline where search terms are generated using GPT-4 by a significant margin. Recall that we test our methods using Retrieval@k for $k \in 35, 40, 45$ to take into account the fact that this allows us to demonstrate that our retrieval mechanism can effectively simplify Task 3, which requires retrieval from the entire corpus, into Task 2, which involves building entailment trees based on relevant facts and a few distractors. 

While our results are not perfect, frame semantic indexing and retrieval has one significant advantage. It is the fact that each stage of the process can be improved by fine-tuning LLMs for the specific purpose. In addition, the transparent nature of this process, which outputs frames at each stage, allows for the analysis and `debugging' of each stage. 

\section{Conclusions and Future Work}
This work presents a novel mechanism for the indexing and retrieval of facts relevant for answering specific questions with the purpose of mitigating factual hallucinations in LLMs. This work demonstrates the feasibility and effectiveness of this method in both retrieval and the automatic generation of frames which, when scaled to multiple tasks, has the potential to provide data-driven insights to the theory of Frame Semantics. 

We believe that this work provides a template for effectively integrating cognitive linguistics and LLM research, benefiting both fields. In future work, we intend to first create models that are specifically fine-tuned to perform each of the tasks described above, specifically frame generation, frame identification and frame relation identification. In addition, we indent to extend this work to multiple tasks so as to establish its effectiveness. 

\section*{Limitations}

This work presents a novel method of indexing and retrieval of facts specifically for the purpose of ensuring that LLMs reason over retrieved, and therefore correct facts, thereby mitigating factual hallucinations in LLMs. Our experiments are based on a single task in a specific domain. As a proof of concept of a novel method that is based on cognitive linguistic theory, these experiments are effective in showcasing the feasibility of this method. However, demonstrating the effectiveness of this method on multiple tasks is required for a more rigorous test, which we leave to future work.  Additionally, our experiments, however, do not extend to testing LLMs for reduced hallucinations -- prior work implies that improved retrieval will indeed lead to reduced hallucinations, but it is left to future work to rigorously test this. 

\bibliography{custom}
\bibliographystyle{acl_natbib}




\end{document}